\documentclass{bmvc2k}

\usepackage{amsmath}
\usepackage{algorithm}
\usepackage{algorithmic}
\usepackage{mathtools}
\usepackage{xfrac}
\usepackage{graphicx}

\DeclarePairedDelimiter{\ceil}{\lceil}{\rceil}

\title{ANSAC: Adaptive Non-Minimal \\ Sample and Consensus}

\addauthor{Victor Fragoso}{victor.fragoso@mail.wvu.edu}{1}
\addauthor{Christopher Sweeney}{csweeney@cs.washington.edu}{2}
\addauthor{Pradeep Sen}{psen@ece.ucsb.edu}{3}
\addauthor{Matthew Turk}{mturk@cs.ucsb.edu}{3}

\addinstitution{
 West Virginia University\\
 Morgantown, WV, USA
}
\addinstitution{
 University of Washington\\
 Seattle, WA, USA
}
\addinstitution{
 University of California\\
 Santa Barbara, CA, USA
}

\runninghead{Fragoso, et al.}{ANSAC: Adaptive Non-Minimal Sample and Consensus}

\def\eg{\emph{e.g}\bmvaOneDot}
\def\ie{\emph{i.e}\bmvaOneDot}

\def\etal{\emph{et al}\bmvaOneDot}

\begin{document}

\maketitle

\begin{abstract}
\vspace{-2mm}
While RANSAC-based methods are robust to incorrect image correspondences (outliers), their hypothesis generators are not robust to correct image correspondences (inliers) with positional error (noise). This slows down their convergence because hypotheses drawn from a minimal set of noisy inliers can deviate significantly from the optimal model. This work addresses this problem by introducing ANSAC, a RANSAC-based estimator that accounts for noise by adaptively using more than the minimal number of correspondences required to generate a hypothesis. ANSAC estimates the inlier ratio (the fraction of correct correspondences) of several ranked subsets of candidate correspondences and generates hypotheses from them. Its hypothesis-generation mechanism prioritizes the use of subsets with high inlier ratio to generate high-quality hypotheses. ANSAC uses an early termination criterion that keeps track of the inlier ratio history and terminates when it has not changed significantly for a period of time. The experiments show that ANSAC finds good homography and fundamental matrix estimates in a few iterations, consistently outperforming state-of-the-art methods.
\end{abstract}
\vspace{-6mm}
\section{Introduction}
\label{sec:introduction}
\vspace{-2mm}

The robust estimation of geometrical models from pairs of images (\eg, homography, essential and fundamental matrices) is critical to several computer vision applications such as structure from motion, image-based localization, and panorama stitching~\cite{agarwal2011, brown2005multi, crandall2013, frahm2010building, li2012, lou2012, sattler2011, zhang2006}. Several applications use robust estimators to compute these transformations from candidate image correspondences that contain outliers (\ie, incorrect image correspondences). While RANSAC-based estimators are robust to outliers, they lack a hypothesis generator that can produce an accurate model from a minimal set of noisy inliers (\ie, correct image correspondences with positional error). Since existing hypothesis generators are sensitive to the noise in the data, the generated hypotheses can deviate significantly from the optimal model, yielding a delay in the estimator's convergence.



Although there exist hypothesis-generation improvements~\cite{brahmachari2013hop, chum2005matching, fragoso2013evsac, fragoso2013swigs, goshen2008balanced, tordoff2005guided} that accelerate the convergence of RANSAC-based estimations, there is no previous work addressing the noise in the correspondences {\it at the hypothesis-generation stage}. The effect of noisy inliers in a hypothesis generated from a minimal sample (a set with the least amount of correspondences to generate a hypothesis) can be reduced by using more samples (non-minimal). This is because more inliers provide additional constraints that help the estimator to produce a more accurate model. However, using a non-minimal sample to generate a hypothesis increases the chances of adding an outlier, which can lead to wrong hypotheses. For this reason, existing methods~\cite{chum2003locally, lebeda2012fixing, raguram2008comparative} that tackle the effect of noisy inliers operate as a refinement stage rather than operating directly in the hypothesis generation phase. While these methods reduce the effect of noise in hypotheses, they can increase the computational overhead since they run an additional inner-refinement stage in the hypothesis-and-test loop of a RANSAC-based estimator.

To reduce the noise effect on hypotheses produced in the hypothesis-generation phase, this work presents a novel Adaptive Non-Minimal Sample and Consensus (ANSAC) robust estimator. Unlike previous work~\cite{brahmachari2013hop, chum2003locally, chum2005matching, fragoso2013evsac, fragoso2013swigs, goshen2008balanced, lebeda2012fixing, raguram2008comparative, tordoff2005guided}, ANSAC produces hypotheses from non-minimal sample sets in the hypothesis generation stage of a RANSAC-based estimator. The homography and fundamental matrix estimation experiments show that ANSAC returns a good estimate consistently and quickly.
\vspace{-5mm}
\subsection{Related Work}
\label{sec:rel_work}
\vspace{-2mm}

The key element of existing methods that accelerate the estimation process is their ability to assess the correctness of every candidate correspondence. Different approaches compute this correctness by means of probabilities, rankings, and heuristics. The probability-based methods (\eg, Guided-MLESAC~\cite{tordoff2005guided}, BEEM~\cite{goshen2008balanced}, BLOGS~\cite{brahmachari2013hop}, and EVSAC~\cite{fragoso2013evsac}) aim to calculate the probability of correctness of each candidate correspondence using the matching scores. They use then the computed probabilities to form a discrete distribution over the candidate correspondences to sample and generate hypotheses.


Instead of randomly sampling candidate correspondences from a probability distribution, PROSAC~\cite{chum2005matching} creates an initial subset with candidate correspondences ranked by quality. PROSAC generates a hypothesis by randomly sampling a minimal number of candidate correspondences from this subset. PROSAC iteratively expands the subset by including lower ranked candidate correspondences and keeps sampling a minimal number of candidate correspondences from this subset until convergence.

While previous methods accelerate the estimation process, they still use minimal samples, which are sensitive to noise. To alleviate this issue, Chum~\etal~\cite{chum2003locally, lebeda2012fixing} proposed LO-RANSAC, an estimator that adds a refinement stage after finding a good hypothesis. This refinement stage is another RANSAC process in which non-minimal samples are generated from the inliers supporting the new best hypothesis.

In contrast to the previous methods, ANSAC uses subsets with a high predicted inlier ratio to draw {\it non-minimal samples} and generate hypotheses that account for noise at the {\it hypothesis-generation phase}. Thus, ANSAC does not require an inner-RANSAC refinement process as in LO-RANSAC, which can be expensive~\cite{lebeda2012fixing}. Consequently, it avoids the extra computational overhead that an inner refinement process incurs. The key element in ANSAC is to predict the inlier ratio of the subsets to determine the size of a non-minimal sample.

\vspace{-6mm}
\section{ANSAC}
\label{sec:description}
\vspace{-3mm}

\begin{figure*}[t]
    \centering
    \includegraphics[width=\textwidth]{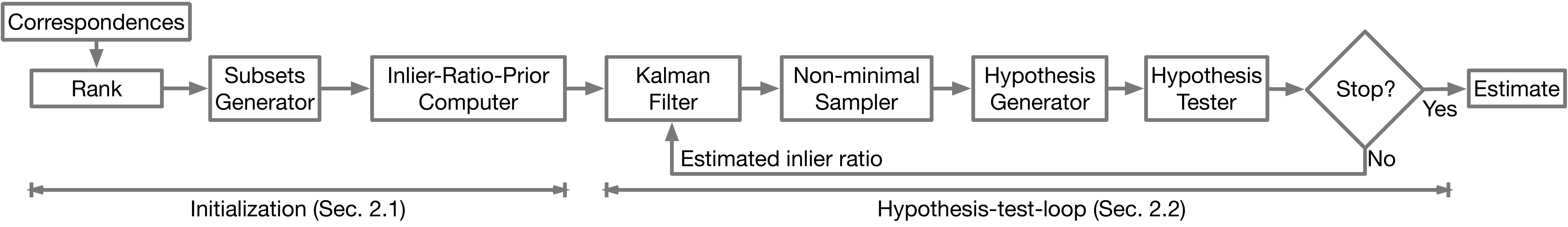}
    \vspace{-8mm}
    \caption{\small{Overview of the pipeline in ANSAC. First, it ranks the correspondences by a correctness quality value. Then, ANSAC builds subsets, and estimates the inlier-ratio prior using a correctness quality value for every subset. Then, the hypothesis-and-test loop uses the subsets to draw hypotheses and tests them to produce a final estimate. ANSAC uses a Kalman filter to refine the inlier-ratio estimate of the current subset from which it generates hypotheses. To do so, the filter combines the inlier ratio prior and the one calculated by testing hypotheses. The non-minimal sampler uses the refined inlier ratio to generate a non-minimal or minimal sample. ANSAC stops when a termination criterion is met.}}
    \label{fig:overview}
    \vspace{-4mm}
\end{figure*}

The goal of ANSAC is to increase the likelihood of finding an accurate hypothesis in the early iterations by generating hypotheses from outlier-free non-minimal samples. To this end, ANSAC prioritizes the use of subsets with high estimated inlier ratios to generate hypotheses. The inlier ratio can be considered as the probability of selecting a correct correspondence from a set at random, since it is the fraction of correct correspondences in a given set. As such, ANSAC uses the inlier ratio to assess the odds of producing an outlier-free non-minimal sample, since it measures the purity of a set. The estimation pipeline of ANSAC (illustrated in Fig.~\ref{fig:overview}) has two main stages: initialization and hypothesis-and-test loop.

\vspace{-4mm}
\subsection{Initialization}
\label{sec:initialization}
\vspace{-2mm}
The main goal of the initialization is to compute subsets of correspondences and their estimated inlier ratio priors. This stage has three main components: correspondence ranking, subset computation, and inlier ratio prior calculation. Given the candidate image correspondences, the initialization stage outputs the ranked subsets $\left\{ \mathcal{S}_i \right\}_{i=1}^{n - m + 1}$ and their initial inlier ratio estimates/priors $\left\{ \varepsilon_i^0 \right\}_{i=1}^{n - m + 1}$, where $n$ is the total number of candidate correspondences, and $m$ is the minimal number of correspondences to generate a hypothesis.

\vspace{-5mm}
\subsubsection{Ranking Correspondences and Subset Generation}
\vspace{-2mm}
The main goal of this stage is to rank high the correspondences that are likely to be correct. This is because ANSAC aims to generate hypotheses from outlier-free samples as early as possible. Thus, ANSAC first calculates their correctness or quality value (\eg, using SIFT ratio~\cite{lowe2004distinctive} or Meta-Recognition Rayleigh (MRR)~\cite{fragoso2013swigs}) and ranks the correspondences. Then, ANSAC builds an initial subset with the minimal number of best-ranked correspondences to draw a hypothesis. This initial subset is the base to generate the remaining subsets. ANSAC generates each subsequent subset by adding the next ranked correspondence to the previously generated subset. In this way, ANSAC keeps growing the most recently generated subset until it uses all the correspondences. The initialization stage computes the $i$-th subset $\mathcal{S}_i$ efficiently by including all the top $j$ ranked correspondences. Mathematically, the $i$-th subset is $\mathcal{S}_i = \left\{ ( \mathbf{x} \leftrightarrow \mathbf{x}^{\prime} )_j : 1 \leq j < m + i, 1 \leq i \leq n - m + 1\right\}$, where $\left( \mathbf{x} \leftrightarrow \mathbf{x}^{\prime} \right)_j$ is the $j$-th ranked candidate correspondence.

\vspace{-6mm}
\subsubsection{Computing the Inlier Ratio Prior}
\label{sec:inlier_ratio_prior}
\vspace{-2mm}
The last step in the initialization stage is the calculation of the inlier ratio prior for every subset. These inlier ratio priors are key to initialize a Kalman filter that will refine the inlier ratio estimate of every subset over time; see Sec.~\ref{sec:kalman_filter}.

The inlier ratio $\varepsilon_i$ of $\mathcal{S}_i$ is defined as the probability of drawing a correct correspondence from it, \ie, $\varepsilon_i = \Pr\left({y = +1 ~|~  \mathcal{S} = \mathcal{S}_i}\right)$, where $y \in \{+1, -1\}$ is a random variable indicating correct or incorrect correspondence, respectively. This probability can be computed as follows: $\varepsilon_i = \frac{1}{|\mathcal{S}_i|} \sum_{j=1}^{|\mathcal{S}_i|} I\left(y_j = +1 \right) \label{eq:inlier_ratio_prior}$, where $I(e)$ is the indicator function which returns 1 when $e$ is true, and 0 otherwise. The ideal indicator function is an oracle that identifies correct correspondences, which is challenging to obtain. Nevertheless, it can be approximated by mapping the correctness or ranking quality values to the $[0, 1]$ range. For instance, using MRR~\cite{fragoso2013swigs} probabilities or mapping Lowe's ratio to the $[0, 1]$ range using a radial basis function. The initialization stage computes the inlier ratio prior $\varepsilon_i^0$ for every subset $\mathcal{S}_i$ using this approximation to the indicator function $\hat{I}$.

\vspace{-5mm}
\subsection{Hypothesis-and-test Loop}
\vspace{-2mm}
The goal of this stage is to robustly estimate a model from the correspondences as quickly as possible. To this end, this stage iteratively draws non-minimal samples when possible from the ranked subsets $\{ \mathcal{S}_i \}_{i=1}^{n-m+1}$ and estimates their inlier ratios $\{ \varepsilon_i \}_{i=1}^{n-m+1}$. Similar to PROSAC~\cite{chum2005matching}, ANSAC progressively utilizes all the subsets according to the rank in order to generate hypotheses. However, unlike PROSAC, which generates hypotheses from minimal samples, ANSAC estimates the subsets' inlier ratios to decide the size of the non-minimal sample for hypothesis generation. Thus, since ANSAC assumes that the subsets at the top of the ranking are less contaminated it will likely use most of the candidate correspondences and generate hypotheses from non-minimal samples when possible. As ANSAC goes through the lowest-ranked subsets, it will likely use minimal samples to generate hypotheses. Similar to PROSAC, ANSAC uses a new subset once it reached the maximum number $M$ of hypotheses it can generate from the current subset. However, unlike PROSAC, which uses a recursive formula to determine the maximum number $M$ of hypotheses to draw from a given subset, ANSAC uses the estimated inlier ratio of a given subset to calculate $M$. 

\begin{algorithm}[t]
\caption{ANSAC}
\label{alg:ansac}
\footnotesize{
\begin{algorithmic}[1]
\REQUIRE{The correspondences $\mathcal{C} \leftarrow \{\left(\mathbf{x} \leftrightarrow \mathbf{x}^{\prime}\right)_j \}_{j=1}^n$}
\ENSURE{Estimate $H^{\star}$}
\STATE{$ \left\{ \left( \mathcal{S}_i, \varepsilon_i^0 \right) \right\}_{i=1}^{n-m+1} \leftarrow$ Initialize($\mathcal{C}$)}
\STATE{Initialize: $s \leftarrow 1$, $\varepsilon_s \leftarrow \varepsilon_s^0$, $\varepsilon_s^{\prime} \leftarrow \varepsilon_s$, $\varepsilon \leftarrow 0$, $\varepsilon_H \leftarrow 0$, $L \leftarrow \infty$~\text{// See Section~\ref{sec:initialization}.}}
\FOR{iteration = 1 \TO $L$}
\STATE{$\left( \mathcal{X}, M \right) \leftarrow$ GenerateSample($\mathcal{S}_s$, $\varepsilon_s$)~~~\text{// Each subset generates at most $M$ hypotheses according to Eq.~\eqref{eq:max_num_iterations_subset}.}}
\STATE{$H \leftarrow$ GenerateHypothesis($\mathcal{X}$)}
\STATE{$(\varepsilon_H, \varepsilon_{s,H}^{\prime}) \leftarrow$ TestHypothesis($H$)}
\IF{$\varepsilon_H > \varepsilon$~~~\text{// The new hypothesis has more inliers than the current best hypothesis.}}
\STATE{$H^{\star} \leftarrow H$, $\varepsilon \leftarrow \varepsilon_H$, $\varepsilon_s^{\prime} \leftarrow \varepsilon_{s,H}^{\prime}$}
\STATE{$L \leftarrow$ UpdateMaxIterations($\varepsilon$)~~~\text{// See Sec.~\ref{sec:termination}.}}
\ENDIF
\STATE{$s \leftarrow$ UpdateSubset(iteration, $M$)~~~\text{// $s \leftarrow s + 1$ when $M$ hypotheses were generated, else $s \leftarrow s $.}}
\STATE{$\varepsilon_s \leftarrow$ RefineInlierRatio($\mathcal{S}_s$, $\varepsilon_s^{\prime}$)}
\IF{Terminate($\varepsilon$)}
\STATE{Terminate loop}
\ENDIF
\ENDFOR
\end{algorithmic}
}
\end{algorithm}

Specifically, ANSAC operates as detailed in Alg.~\ref{alg:ansac}. Given the input subsets $\{ \mathcal{S}_i \}_{i=1}^{n-m+1}$ and their inlier ratio priors $\{ \varepsilon_i^0 \}_{i=1}^{n-m+1}$, ANSAC sets its current-subset index $s$ pointing to the first subset, \ie, $s \leftarrow 1$. Subsequently, it generates a sample $\mathcal{X}$ from the current subset $\mathcal{S}_s$ in step 4. However, unlike existing estimators that always draw minimal samples, ANSAC uses the current inlier ratio estimate $\varepsilon_s$ to determine the size of the sample; see Sec.~\ref{sec:sampler}. Given the non-minimal sample, ANSAC generates a hypothesis $H$ in step 5. In step 6, the hypothesis tester identifies two sets of correspondences: 1) those that support the generated hypothesis $H$ across all the input correspondences to compute its overall inlier ratio $\varepsilon_H$; and those in the current subset to calculate the subset's inlier ratio $\varepsilon^{\prime}_{s,H}$. When the inlier ratio $\varepsilon_H$ is larger than the inlier ratio $\varepsilon$ of the current best hypothesis, then ANSAC updates several variables (steps 7-10): the current best hypothesis $H^\star \leftarrow H$, the overall inlier ratio $\varepsilon \leftarrow \varepsilon_H$, the ``observed'' inlier ratio $\varepsilon_s^{\prime} \leftarrow \varepsilon^{\prime}_{s,H}$ (the fraction of correspondences from the current subset $\mathcal{S}_s$ supporting the new best hypothesis $H^\star$), and the maximum number $M$ of hypotheses to draw from $\mathcal{S}_s$. In step 11, ANSAC uses a new subset by increasing $s \leftarrow s + 1$ if it has generated $M$ hypotheses from the current subset. Then, a Kalman filter refines the inlier ratio estimate $\varepsilon_s$ of the current subset in step 12; see Sec.~\ref{sec:kalman_filter}. Lastly, ANSAC checks if the current best hypothesis $H^\star$ satisfies a termination criteria and stops if satisfied (steps 13-15). 

\vspace{-4mm}
\subsubsection{Estimating the Inlier Ratio with a Kalman Filter}
\label{sec:kalman_filter}
\vspace{-2mm}

The Kalman filter~\cite{fox2003bayesian, thrun2005probabilistic} iteratively refines the estimate of the inlier ratio of the current subset $\varepsilon_s$. It does so by combining the inlier ratio prior $\varepsilon_s^0$ and the observed inlier ratio $\varepsilon_s^{\prime}$. This is a crucial component since the inlier ratio is the value that determines the size of the sample and the progressive use of subsets. To use this filter, ANSAC assumes that the inlier ratio varies linearly as a function of the ranked subsets and it has Gaussian noise.

The prediction step of the Kalman filter estimates the inlier ratio $\varepsilon_s^{k}$ at time $k$ given the inlier ratio $\varepsilon_s^{k-1}$ at previous time and the inlier ratio prior $\varepsilon_s^0$. The filter uses the following linear inlier ratio ``motion'' model:
\vspace{-2mm}
\begin{align}
    \varepsilon_{k|k-1} &= \alpha_k \varepsilon_s^{k-1} + \beta_k \varepsilon_s^0  \label{eq:kf_motion_model} \\
    \sigma_{k|k-1}^2 &= \alpha_k^2 \sigma_{s, k-1}^2 + \sigma_p^2,
\end{align}
where $\varepsilon_{k|k-1}$ is the predicted inlier ratio, $\sigma_{k|k-1}$ is the standard deviation for the predicted inlier ratio, $\alpha_k$ and $\beta_k$ are weights that control the contribution of their respective inlier ratio terms, $\sigma_{s,k-1}$ is the standard deviation of the previous estimate $\varepsilon_s^{k-1}$, and $\sigma_p$ is a standard deviation parameter that enforces stochastic diffusion. The weights $\alpha_k$ and $\beta_k$ must satisfy $1 = \alpha_k + \beta_k$ since the inlier ratios range between $[0, 1]$.

To initialize the filter, ANSAC sets $\varepsilon_s^{k-1} \leftarrow \varepsilon_s^0$ when it is the first iteration. However, when ANSAC uses a new subset, then ANSAC sets $\varepsilon_s^{k-1} \leftarrow \varepsilon_{s-1}$, where $\varepsilon_{s-1}$ is the estimate of the previous set. This is a good initial value since subsets only differ by one sample. ANSAC uses the following function to set $\alpha_k = \alpha^{I(s \leftarrow s + 1)}$, where $I(s \leftarrow s + 1)$ returns $1$ when a new current subset is used, and $0$ otherwise; and the parameter $\alpha$ satisfies $0 \leq \alpha \leq 1$. This function forces the filter to use the inlier ratio prior only when ANSAC uses a new subset. This is because the inlier ratio of a subset can be refined over multiple iterations and the prior must contribute once to the refinement. 



The update step aims at refining the output of the prediction step. To this end, it combines the observed inlier ratio $\varepsilon_s^{\prime}$, which is obtained by testing the best current hypothesis on the current subset. The update step performs the following computations:
\begin{align}
    g &= \frac{\sigma^2_{k|k-1}}{\sigma^2_{k|k-1} + \sigma^2_u} \\
    \varepsilon_s^{k} & = \varepsilon_{k|k-1} + g (\varepsilon_s^{\prime} - \varepsilon_{k|k-1}) \\
    \sigma^2_{s,k} & = (1 - g) \sigma^2_{k|k-1},
\end{align}
where $\sigma^2_u$ is a parameter enforcing stochastic diffusion, $g$ is the Kalman gain, $\varepsilon_s^{k}$ is the refined inlier ratio for the current subset at time $k$, and $\sigma^2_{s,k}$ is its standard deviation. At the end of this stage, ANSAC updates the inlier ratio of the current subset $\varepsilon_s = \varepsilon_s^k$.

\vspace{-4mm}
\subsubsection{Adaptive Non-minimal Sampling}
\label{sec:sampler}
\vspace{-2mm}
Unlike many RANSAC variants (\eg,~\cite{brahmachari2013hop, chum2005matching, fischler1981random, fragoso2013swigs, fragoso2013evsac, goshen2008balanced, raguram2008comparative, raguram2011recon}) which use minimal samples to generate hypotheses, ANSAC computes the size $q$ of a sample as follows:
\begin{equation}
    q(\varepsilon_s) = \left( 1 - \lambda(\varepsilon_s) \right) q_{\text{\small min}} + \lambda(\varepsilon_s) q_{\text{\small max}},
    \label{eq:adaptive_sample_size}
\end{equation}
where $q_{\text{\small min}} = m$ is the minimum sample size to generate a hypothesis; $q_{\text{\small max}}$ is the maximum sample size, which is upper-bounded by the size of the current subset $\mathcal{S}_s$; and
\begin{equation}
    \lambda(\varepsilon_s; \omega, \mu) = \left( 1 + \exp{\left(-\omega\left(\varepsilon_s - \mu\right)\right)}\right)^{-1},
    \label{eq:logistic_fn}
\end{equation}
is a logistic curve with parameters $\omega$ (steepness) and $\mu$ (inflection point). The value constraints for these parameters are: $0 < \omega$ and $0 < \mu < 1$.

Eq.\eqref{eq:adaptive_sample_size} selects a sample size by linearly interpolating between $q_{\text{\small min}}$ and $q_{\text{\small max}}$. The sample size depends directly on measuring the likelihood of not including outliers, which is measured by $\lambda$ (see Eq.$\eqref{eq:logistic_fn}$). When the likelihood is high ($> 0.5$), ANSAC produces a non-minimal sample. Otherwise, it produces minimal samples.  ANSAC evaluates this function at every iteration since $\varepsilon_s$ evolves over time.

\vspace{-5mm}
\subsubsection{Maximum Number of Hypotheses per Subset}
\vspace{-2mm}
ANSAC generates a maximum number $M$ of hypotheses from the current subset $\mathcal{S}_s$. ANSAC considers the worst case scenario where the inlier ratio of the current subset is low and thus the sample size is the minimal. Thus, it calculates $M$ using the following formula~\cite{fischler1981random, raguram2013usac}:
\begin{equation}
    M = \ceil[\Bigg]{\frac{\log\left(1 - \nu \right)}{\log\left( 1 - p \right)}},
    \label{eq:max_num_iterations_subset}
\end{equation}
where $\nu$  is the probability of producing a good hypothesis, and $p = (\varepsilon_s^{\prime})^{q_{\min}} $ is the probability of a minimal sample being outlier-free. ANSAC adapts $M$ as a function of $\varepsilon_s^{\prime}$ every iteration. Eq.~\eqref{eq:max_num_iterations_subset} provides the least amount of hypotheses drawn from minimal samples that an estimator needs to test to achieve a confidence $\nu$~\cite{fischler1981random, raguram2013usac}. The progressive use of the subsets in ANSAC depends directly on the estimated inlier ratio of each subset, unlike PROSAC~\cite{chum2005matching} which uses a growth function that depends on the number of samples drawn by each subset.

\vspace{-5mm}
\subsubsection{Early Termination}
\label{sec:termination}
\vspace{-2mm}
The classical termination criterion of a RANSAC-based estimator calculates the maximum number of iterations adaptively using Eq.~\eqref{eq:max_num_iterations_subset}, since one iteration generates and tests a hypothesis. Even though that an estimator finds a good hypothesis in the early iterations, the estimator still iterates until it reaches the maximum number of iterations calculated with Eq.~\eqref{eq:max_num_iterations_subset}. In this case, the iterations after finding a good hypothesis are unnecessary.

To alleviate this problem, many approaches aim to detect if a hypothesis is likely to be bad~\cite{capel2005effective, chum2002randomized, matas2005randomized} to skip the iteration, and thus avoid testing the hypothesis. Unlike these methods, ANSAC aims to detect if the estimator is unlikely to find a significantly better hypothesis than the current best one, and terminate the loop. To this end, ANSAC uses the history of the overall inlier ratio $\varepsilon$ throughout of the iterations. Since ANSAC uses ranked correspondences to generate hypotheses, it is expected to see a rapid increase of the overall inlier ratio $\varepsilon$ in the early iterations, and a slow or none increase in the later iterations. The proposed termination criterion detects when $\varepsilon$ has reached the low-or-none increase zone.

The proposed early termination criterion analyzes the latest $T$ increments of the overall inlier ratio $\left\{ \delta_{\varepsilon}(x_t) \right\}_{t=1}^T$, where $\delta_{\varepsilon}(l) = \varepsilon(l) - \varepsilon(l - 1)$ and $\varepsilon(l)$ is the overall inlier ratio at the $l$-th iteration. ANSAC decides to stop iterating if their average variation  $\bar{\delta}_{\varepsilon}$ is below a certain threshold $\gamma$. The value of $T$ is bounded by the overall maximum number of iterations $L$, which is calculated using $p = \varepsilon^m$ and Eq.~\eqref{eq:max_num_iterations_subset}. Thus, $T$ has to change adaptively as well. For this reason, ANSAC uses $T = \tau L$, where $0 < \tau < 1$, to set the value for $T$.

\vspace{-3mm}
\section{Experiments}
\label{sec:experiments}
\vspace{-2mm}

This section presents a series of experiments, each demonstrating the effectiveness and efficiency of the proposed components in ANSAC. We implemented ANSAC in C++ and use the robust estimation API from the TheiaSfM library~\cite{sweeney2015theia}\footnote{Source code available at~\url{http://vfragoso.com/}}. We used the implementations of Root-SIFT~\cite{arandjelovic2012three} and CasHash~\cite{cheng2014fast} in TheiaSfM to match features and compute symmetric correspondences: common image correspondences obtained by matching image one as a reference and image two as a query and vice versa.

\vspace{-5mm}
\paragraph{Datasets.} We used the Oxford affine-covariant-regions dataset~\cite{mikolajczyk2005comparison} and USAC homography-image pairs~\cite{raguram2013usac} for homography estimation. In addition, we used the USAC fundamental-matrix-image pairs and Strecha dataset~\cite{strecha2004wide, strecha2006combined} for fundamental matrix estimation. The Oxford and Strecha dataset provide information to calculate ground truth. To get a ground truth for USAC dataset, we used the same procedure used by Raguram~\etal~\cite{raguram2013usac}, which estimates the models by running RANSAC $10^7$ times, followed by a visual inspection of the image correspondences to ensure the models did not include outliers. See supplemental material for a visualization of the image pairs from these datasets.

\vspace{-5mm}
\subsection{Kalman Filter for Inlier Ratio Estimation}
\vspace{-2mm}

\begin{figure}[t]
\centering
\includegraphics[width=\textwidth]{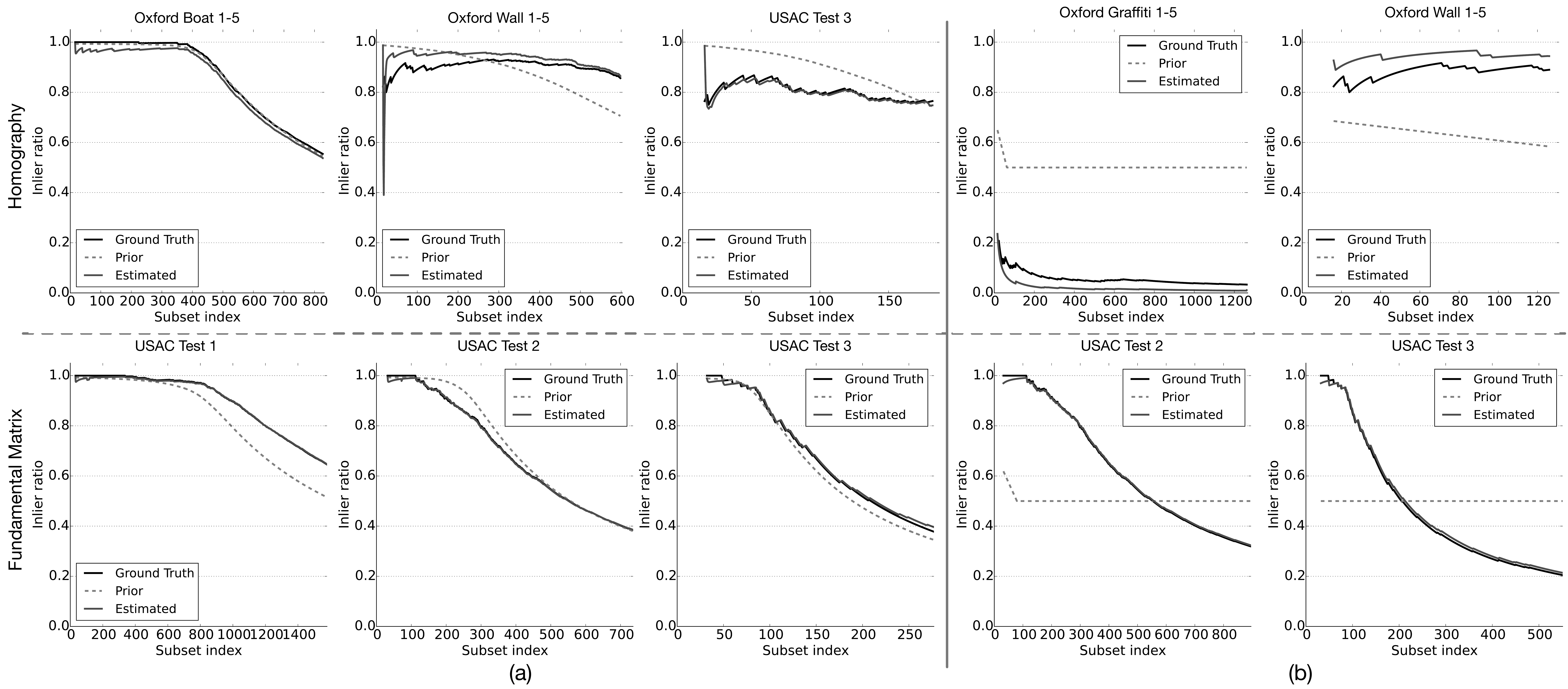}
\vspace{-8mm}
\caption{\small{{\bf(a)} Accuracy of the subset-inlier-ratio estimation for Homography (top row) and Fundamental matrix (bottom row) models. The filter tends to provide a close estimation (gray curve) to the ground truth (black curve). {\bf(b)} Accuracy of the estimation using a bad inlier-ratio prior. The filter is robust and tends to provide a close estimation in this scenario.}}
\vspace{-6mm}
\label{fig:kalman_evaluation}
\end{figure}

The goal of this experiment is twofold: 1) measure the subset-inlier-ratio estimation accuracy of the  Kalman filter; and 2) evaluate the robustness of the filter given a poor inlier ratio prior estimate. We used the Lowe's ``radius'' as a correspondence quality: $r^2 = r_1^2 + r_2^2$, where $r_1$ and $r_2$ are the Lowe's ratios obtained by matching image one to image two and viceversa. We used $\sigma_p = \sigma_u = 0.1$ for stochastic diffusion, set $\alpha=0.5$ for the prediction step (see Eq.~\eqref{eq:kf_motion_model}), and removed the early termination criterion for this experiment.

Fig.~\ref{fig:kalman_evaluation} (a) shows the estimation accuracy results for homography and fundamental matrix estimation. It presents the ground truth inlier-ratio (black curve), the estimated inlier-ratio (gray curve), and the inlier ratio prior (dashed) computed from correspondence qualities (see Sec.~\ref{sec:inlier_ratio_prior}) as a function of the sorted subsets. The estimate curve tends to be close to the ground truth even when the inlier-ratio priors are not accurate. This experiment confirms the efficacy of the filter to estimate the inlier-ratio of the subsets in ANSAC. To test the robustness of the filter to bad inlier-ratio priors, we replaced the inlier-ratio priors from the correctness quality values with synthetic estimates that deviate significantly from the ground truth. The results of this experiment are shown in Fig.~\ref{fig:kalman_evaluation} (b). The filter is able to ``track'' the ground truth for both models, even though it used priors that deviated significantly from the ground truth. See supplemental material for larger figures.

\begin{figure}[t]
\centering
\includegraphics[width=\textwidth]{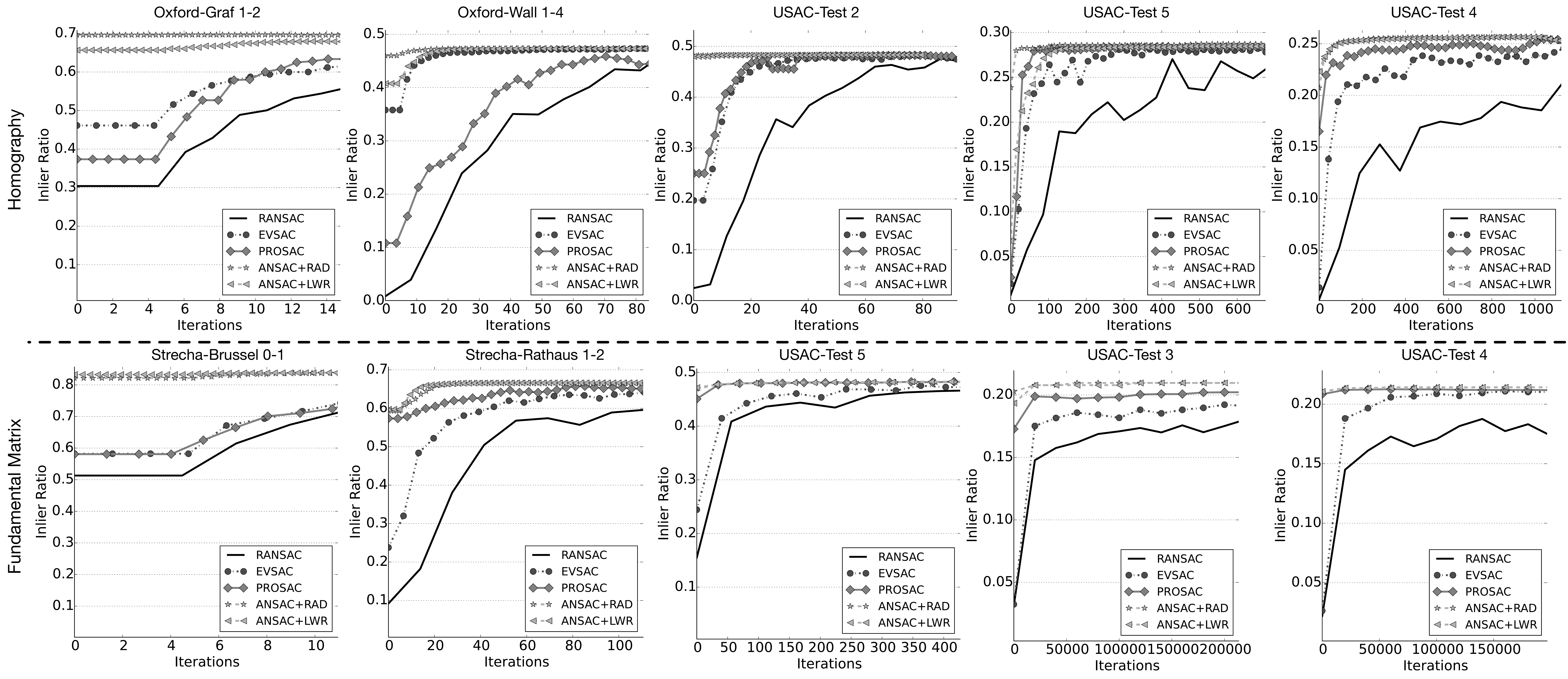}
\vspace{-8.5mm}
\caption{\small{Convergence analysis for homography (top row) and fundamental matrix (bottom row) estimation. ANSAC tends to converge in the earliest iterations, while the competing methods (EVSAC, PROSAC, and RANSAC) require more iterations.}}
\label{fig:convergence}
\vspace{-5mm}
\end{figure}

\vspace{-6mm}
\subsection{Convergence Analysis}
\vspace{-2mm}

The goal of this experiment is to analyze the benefits that the adaptive non-minimal sampler brings to the convergence of ANSAC. We did not use an early termination criterion for this experiment. The experiment was run with the following parameters: $\alpha=0.5$ and $\sigma_p = \sigma_u = 0.1$ for the filter, $\omega=20$ and $\mu=\frac{3}{4}$ which are the parameters of $\lambda$ (see Eq.~\eqref{eq:logistic_fn}), and $q_{\text{max}} = 4 \times q_{\text{min}}$ (see Eq.~\eqref{eq:adaptive_sample_size}). The experiment considered the following competing estimators: RANSAC~\cite{fischler1981random} (the baseline), EVSAC~\cite{fragoso2013evsac}, and PROSAC~\cite{chum2005matching}. We used the TheiaSfM implementations of the aforementioned estimators. The experiment used ANSAC with two different correctness correspondence quality measures: 1) Lowe's radius (ANSAC+RAD), introduced above; and 2) the widely used Lowe's ratio (ANSAC+LWR). Also, the experiment used correspondences ranked using Lowe's ratio for PROSAC, and calculated the correctness probabilities from descriptor distances for EVSAC. A trial in this experiment measured the estimated overall inlier ratio of every estimator as a function of the iterations in their hypothesis-and-test loop. The experiment performed $1000$ trials, and fitted a curve to all its trials for every estimator in order to get a smooth summary.

The top and bottom row in Fig.~\ref{fig:convergence} show the results for homography and fundamental matrix estimation, respectively. The plots are sorted from left-to-right according to their ground truth overall inlier ratio. The plots show that ANSAC tends to converge faster than the competing methods. ANSAC converges faster than the competing methods when the inlier ratio is high ($>0.5$). However, when the inlier ratio is $<0.5$, ANSAC tends to converge faster or comparable than PROSAC. This is because when the overall inlier ratio of an image pair is high, ANSAC uses non-minimal samples which yields a faster convergence than that of the competing methods. On the other hand, when the overall inlier ratio is low, ANSAC tends to use minimal samples more often, which yields a comparable performance than that of PROSAC. See supplemental material for more results about convergence.

\vspace{-6mm}
\subsection{Estimation of Homography and Fundamental Matrix}

\begin{figure}[t]
\centering
\includegraphics[width=\textwidth]{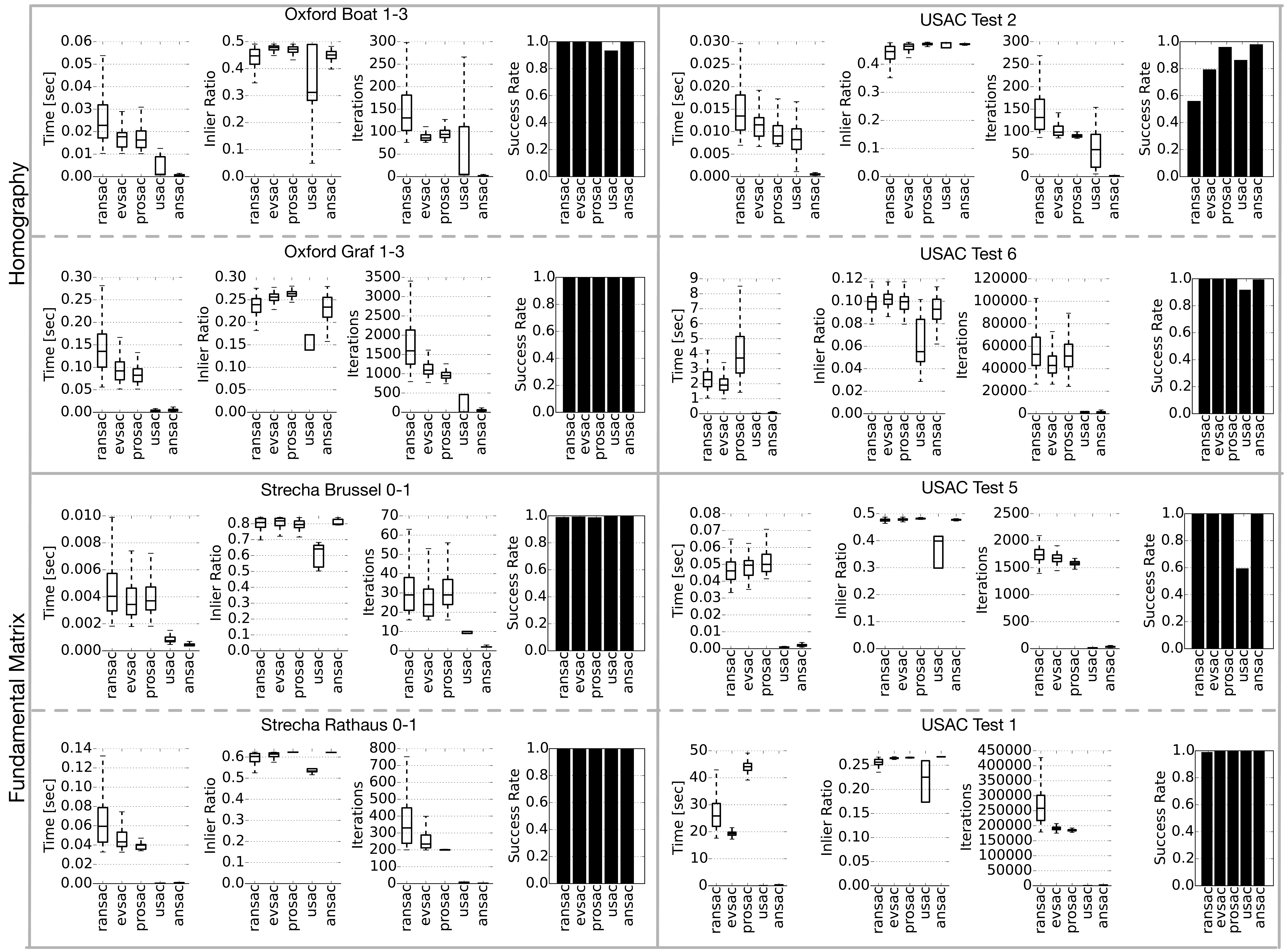}
\vspace{-7mm}
\caption{\small{Box-plots measuring (from left to right) wall clock time, overall inlier ratio, iterations, and success rate of methods estimating homography (top two rows) and fundamental (bottom two rows) matrix models. The plots show that ANSAC and USAC are the fastest; in most cases less than 20 msec. Unlike USAC, ANSAC achieves a similar inlier ratio and success rate that that of RANSAC, PROSAC, and EVSAC for both model estimations.}}
\label{fig:accuracy}
\vspace{-4mm}
\end{figure}

The goal of this last experiment is to measure the speed (wall clock time) of an estimator to converge, the number of iterations, and the success rate and the overall inlier ratio of the estimated model. The experiment consisted of $1000$ trials of each estimator. We measured success rate as the ratio between the number of estimates that are close enough to the ground truth and the total number of trials. This experiment considered RANSAC, PROSAC, EVSAC, and USAC~\cite{raguram2013usac} as part of the competing methods. We used USAC's publicly available C++ source, and enabled PROSAC+SPRT~\cite{matas2005randomized}+LO-RANSAC~\cite{chum2003locally}. We used ANSAC with the Lowe's radius and enabled the termination criterion using $\tau=0.01$. 

The top and bottom two rows in Fig.~\ref{fig:accuracy} present the results for homography and fundamental matrix estimation, respectively. The plots show that ANSAC and USAC were the fastest estimators. ANSAC consistently required less than 20 msec, while USAC presented more time variation in homography estimation (see first row in Fig.~\ref{fig:accuracy}). ANSAC computed estimates whose inlier ratio and success rates are comparable to those of RANSAC, EVSAC, and PROSAC. On the other hand, estimates computed with USAC presented a larger inlier ratio variation. These experiments confirm that ANSAC can accelerate the robust estimation of models using an adaptive non-minimal sampler, while still achieving a comparable or better performance than the state-of-the-art estimators. See the supplemental material for larger plots and extended results on all the experiments.

\section{Conclusion}
\label{sec:conclusions}
\vspace{-3mm}

We have presented ANSAC, an adaptive, non-minimal sample and consensus estimator. Unlike existing estimators, ANSAC adaptively determines the size of a non-minimal sample to generate a hypothesis based on the inlier ratio estimated with a Kalman filter. In contrast to LO-RANSAC methods, which use non-minimal samples in a refinement process, ANSAC uses them in the hypothesis generation phase, avoiding the computational cost of the additional refinement phase. The homography and fundamental matrix estimation experiments demonstrate that ANSAC can converge in the early iterations and perform consistently better or comparable than the state-of-the-art.
\vspace{-4mm}
\paragraph{Acknowledgments.} This work was supported in part by NSF grants IIS-1657179, IIS-1321168, IIS-1619376, and IIS-1423676.

{\small
\bibliography{ansac}
}
\end{document}